\documentclass[11pt]{article}

\usepackage[preprint]{acl}

\usepackage{times}
\usepackage{latexsym}

\usepackage[T1]{fontenc}

\usepackage[utf8]{inputenc}

\usepackage{microtype}

\usepackage{inconsolata}

\usepackage{graphicx}
\usepackage{caption}
\usepackage{subcaption}
\usepackage{amsmath}
\usepackage{booktabs}
\usepackage{color,soul}
\usepackage[table]{xcolor}
\usepackage{pifont}
\usepackage{amssymb}
\usepackage{enumitem}
\usepackage{mathtools}
\usepackage{multirow}
\usepackage{float}
\usepackage{xspace}
\usepackage{tcolorbox}   
\usepackage{wrapfig}
\usepackage{tabularx}
\usepackage{colortbl}
\usepackage{array}
\usepackage[capitalize,noabbrev]{cleveref}

\newcommand{\red}[1]{\textcolor{red}{#1}}
\newcommand{\green}[1]{\textcolor{teal}{#1}}

\newcommand{\headingwithline}[2][10em]{%
    \noindent
    \rule[0.5ex]{\dimexpr\textwidth-#1\relax}{0.5pt}
    \textsc{#2}
    \rule[0.5ex]{\dimexpr\textwidth-#1\relax}{0.5pt}\par%
}

\title{
Inclusion-of-Thoughts: Mitigating Preference Instability \\via Purifying the Decision Space
}

\author{
    \textbf{Mohammad Reza Ghasemi Madani\thanks{Corresponding authors},} \\
    \textbf{Soyeon Caren Han$^*$,}\;
    \textbf{Shuo Yang,}\;
    \textbf{Jey Han Lau}\\
    School of Computing and Information Systems, The University of Melbourne
}

\begin{document}
\maketitle
\begin{abstract}
Multiple-choice questions (MCQs) are widely used to evaluate large language models (LLMs).
However, LLMs remain vulnerable to the presence of plausible distractors. They often divert attention toward irrelevant choices, resulting in unstable oscillation between correct and incorrect answers.
In this paper, we propose Inclusion-of-Thoughts (IoT), a progressive self-filtering strategy that is designed to mitigate this cognitive load (\emph{i.e.}, instability of model preferences under the presence of distractors) and enable the model to focus more effectively on plausible answers. 
Our method operates to reconstruct the MCQ using only plausible option choices, providing a controlled setting for examining comparative judgements and therefore the stability of the model’s internal reasoning under perturbation.
%
Extensive empirical evaluation demonstrates that IoT substantially boosts chain-of-thought performance across a range of arithmetic, commonsense reasoning, and educational benchmarks with {minimal} computational overhead.
\footnote{Code will be released once the paper is accepted.}

\end{abstract}
\section{Introduction}

\begin{quote}
    \textit{``The greatest gift we can offer each other is a second chance.''} --Lloyd D. Newell
\end{quote}

\begin{figure}[t]
    \centering
    \includegraphics[width=1\linewidth]{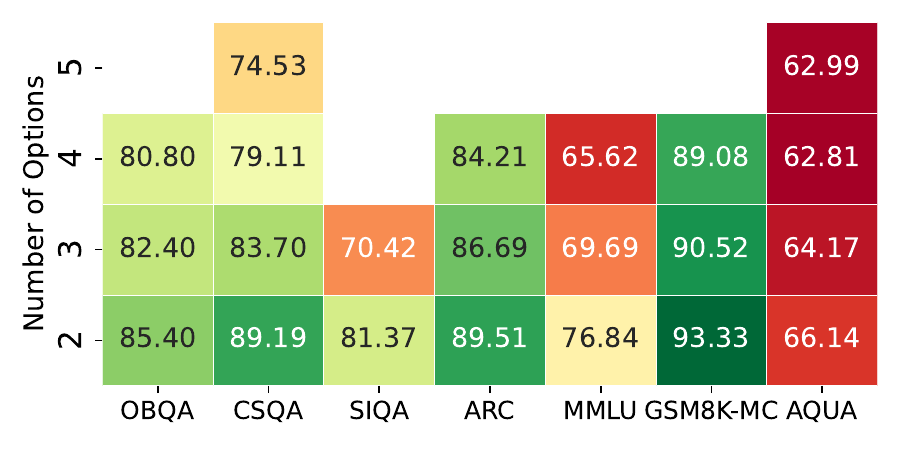}
    \caption{The figure illustrates the effect of options on the accuracy of \texttt{Olmo-2-7B} across different datasets. The y-axis indicates the number of options passed with the question. 
    }
    \label{fig:load}
\end{figure}

%
\citet{wei2023chainofthoughtpromptingelicitsreasoning} propose chain-of-thought (CoT) prompting, where a large language model (LLM) is prompted to generate a series of short sentences that mimic the reasoning process a person might use in solving a task. 
While effective, a number of studies found that when CoT prompting fails, they are sometimes attributed to
LLM's instability in decision making when exposed to multiple plausible distractors~\cite{ma2023poeprocesseliminationmultiple, fu-etal-2025-exclusion, balepur2024s}. 
That is, even though the correct answer has a high probability under the LLM, when presented with more options, the correct answer sees a consistent decrease of probability may result in the model changing its selection and therefore, reduction in the final performance (\cref{fig:load}) --- we call this phenomenon \emph{preference instability}.

\begin{figure}
    \centering
    \includegraphics[width=1\linewidth]{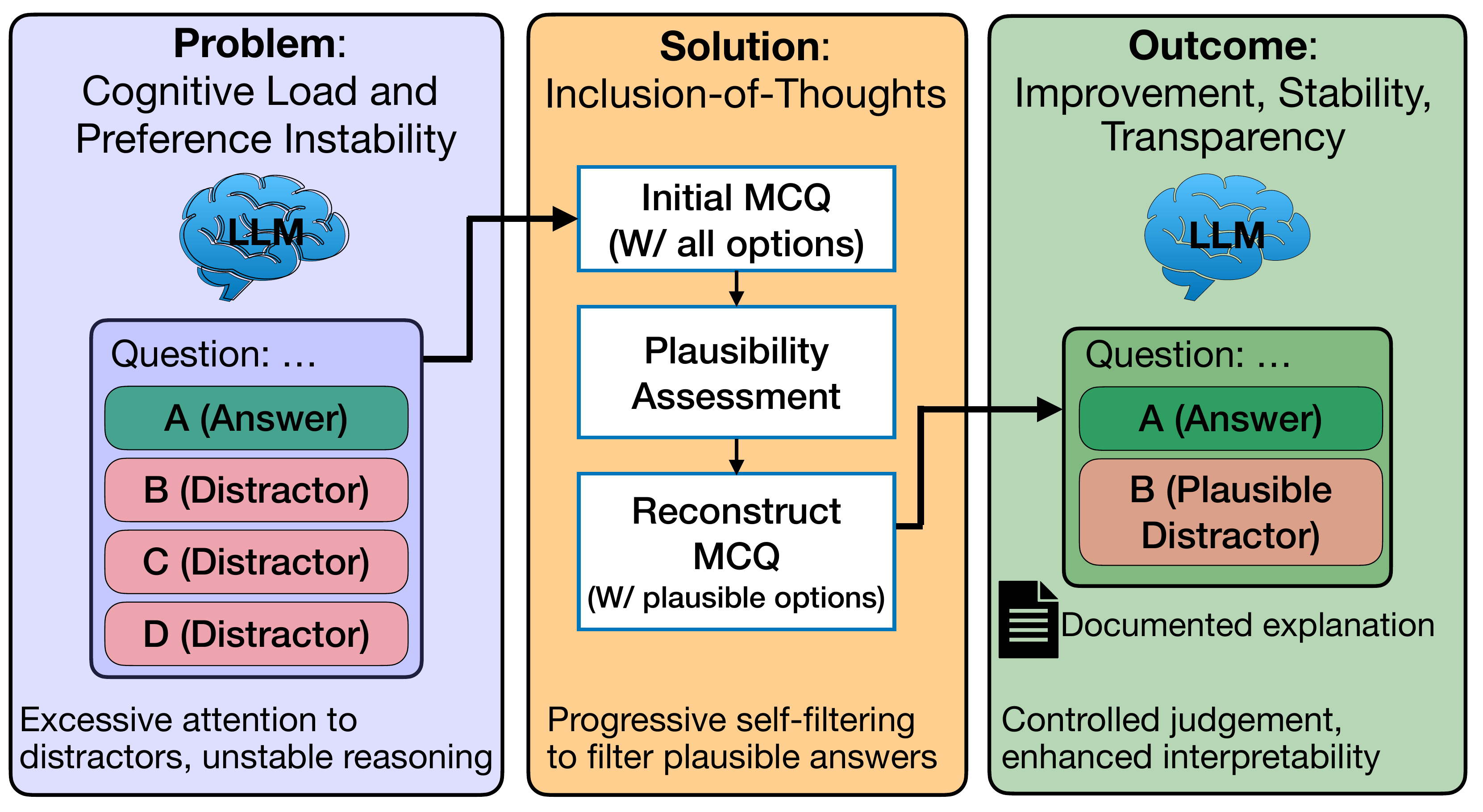}
    \caption{\textbf{Inclusion-of-Thoughts Framework} is a self-filtering strategy that reconstructs multiple-choice questions using only plausible options to mitigate model instability and enhance performance.}
    \label{fig:iot}
\end{figure}

Guided by this insight, we introduce \textbf{Inclusion-of-Thoughts (IoT)}, a novel test-time approach designed to augment the standard CoT process by focusing on LLM's highest-ranked predictions. As illustrated in Figure \ref{fig:iot}, IoT achieves this by strategically \emph{perturbing the option} to elicit and then isolate the model's top two preferences, followed by a final, two-choice judgement.
In contrast to other test-time methods, IoT is computationally lightweight and does not require extensive sampling, complex feedback loops, or tree search. It achieves its performance gain by purifying 
the decision space to focus on the model's top-ranked hypotheses. This makes IoT not only a scalable and cost-effective method for improving reasoning performance; it can also serve as a probing test to understand LLM behaviours.
Another key advantage of IoT is its zero-shot, unsupervised implementation. It works with closed- or open-source LLMs, requiring no additional human annotation, auxiliary models, extra training, or fine-tuning. 

We evaluate IoT on a wide range of arithmetic, commonsense reasoning, and educational tasks over 4 LLMs of varying sizes: \texttt{Llama-3.3-8B}~\cite{llama3modelcard}, \texttt{Olmo-2-7B/13B}~\cite{olmo20242olmo2furious}, and GPT-4o-mini~\cite{openai2024gpt4ocard}. 
Across all LLMs, IoT improves over CoT prompting by a substantial margin across most tasks. In particular,  \texttt{Olmo-2-7b} with IoT achieves surpass CoT across commonsense reasoning tasks (OBQA~\cite{mihaylov2018suitarmorconductelectricity} ($+3.40\%$), CSQA~\cite{talmor2019commonsenseqaquestionansweringchallenge} ($+2.05\%$), GSM8K-MC~\cite{cobbe2021trainingverifierssolvemath} ($+3.95\%$) and educational benchmarks tasks (ARC~\cite{clark2018thinksolvedquestionanswering} ($+3.33\%$) and MMLU~\cite{hendrycks2021measuringmassivemultitasklanguage} ($+1.16\%$)).
To summarise, our contributions are:

\begin{enumerate}[leftmargin=*]
    \item We propose IoT, a lightweight, zero-shot test-time framework that improves MCQ reasoning by isolating an LLM's high probability candidates rather than expanding the reasoning space through sampling or search.

    \item We reveal that errors in MCQ reasoning often arise from preference instability among plausible options rather than missing knowledge or insufficient reasoning steps.

    \item We present a transition-based analysis of LLM predictions across stages, providing diagnostic insight into when and why correct intermediate preferences fail to translate into correct final answers.

    \item Through extensive experiments across commonsense, educational, and mathematical benchmarks, we demonstrate that IoT consistently improves accuracy over strong baselines while incurring minimal computational overhead.
\end{enumerate}

\section{Related Works}

\paragraph{Reasoning with LLMs}

The ability of large language models (LLMs) to generate intermediate steps, a.k.a. Chain-of-Thought (CoT) \citep{wei2023chainofthoughtpromptingelicitsreasoning}, has become a central focus in reasoning research. Several works have explored the internal mechanics of CoT, probing its faithfulness \citep{lyu2023faithful, lanham2023measuring} and its utility in complex planning \citep{saparov2022language}. Further prompt engineering techniques have been developed to enhance this process by encouraging the model to consider multiple viewpoints or generate supporting evidence for competing options, such as Debate Prompting \citep{michael2023debate} and Maieutic Prompting \citep{jung2022maieutic}.
Prior to the rise of general CoT methods, research largely focused on specialized approaches for tasks like mathematical reasoning \citep{andor2019givingbertcalculatorfinding, ran2019numnetmachinereadingcomprehension, geva2020injectingnumericalreasoningskills, piękos2021measuringimprovingbertsmathematical}. In contrast, the introduction of Self-Consistency (SC) by \citet{wang2023selfconsistencyimproveschainthought} provided a more generalized framework, applicable to a wide range of reasoning tasks without requiring additional supervision or fine-tuning, thereby substantially improving the effectiveness of standard CoT.

\paragraph{Test-time Scaling}

A large body of research has shown that LLM reasoning can be dramatically improved via advanced test-time approaches that combine prompting and aggregation. Representative methods include the aforementioned CoT and SC, as well as Tree-of-Thought prompting \citep{yao2023tree}, and Self-Reflection mechanisms \citep{shinn2023reflexion, madaan2023selfrefineiterativerefinementselffeedback}. Furthermore, techniques that leverage backward reasoning have been proposed to verify intermediate steps and improve precision, particularly in mathematical contexts \citep{weng2023large, xue2023rcot, jiang2024forward}.
Orthogonal to search- or verification-heavy test-time scaling, recent analyses have shown that LLM predictions can be highly sensitive to superficial perturbations such as option ordering or minor context changes, indicating unstable preference structures rather than reasoning deficiencies \citep{pezeshkpour2024large}. These findings suggest that improving inference-time robustness may require mechanisms that explicitly probe and stabilize model preferences under controlled perturbations.
While these advanced test-time methods are effective, particularly compared to baseline CoT, they often yield only moderate additional improvement over simpler techniques such as SC alone. Crucially, many of these highly complex approaches, such as backward verification and specific multi-agent setups, have been primarily developed and validated for mathematical tasks, which inherently limits their generalizability to the broader linguistic and commonsense challenges targeted by our framework. Our proposed IoT framework is unique in that it focuses on improving the selection mechanism when there are multiple plausible outputs  without introducing complex, task-specific reasoning steps.
Unlike prior methods that expand or verify reasoning trajectories, IoT treats preference instability itself as the primary failure mode and addresses it through structured option-level perturbation, enabling targeted improvement without increasing reasoning depth or search complexity.

\section{Inclusion-of-Thoughts}
\label{sec:iot}
\begin{figure*}
    \centering
    \includegraphics[width=1\textwidth]{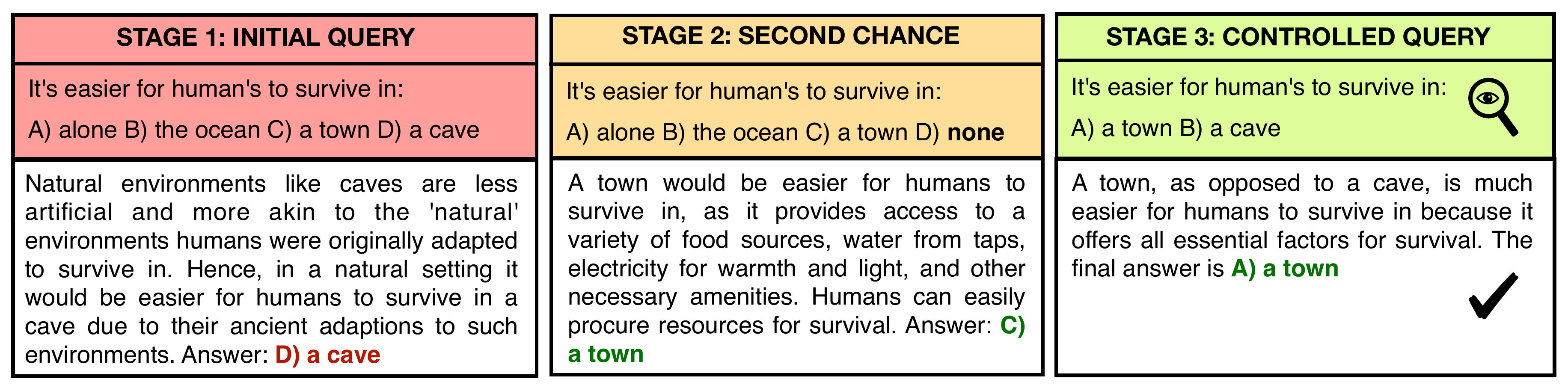}
    \caption{\textbf{Inclusion-of-Thoughts Framework.} The pipeline allows the model to choose up to two options. Then the model looks at its preferences and decides the final answer in isolation.}
    \label{fig:iot-example}
\end{figure*}

\subsection{Methodology}

%
%
IoT operates in three stages:

\paragraph{Stage 1: Initial Preference Elicitation}
Given an MCQ with the original option set $O = \{o_1, \ldots, o_n\}$, the LM
is first prompted using a standard CoT approach to generate its most probable answer, which we denote as the top-choice candidate $o^*_1 \in O$. This initial output reflects the model's raw preference over the full set.


\paragraph{Stage 2: Second Plausibility Assessment}
To probe beyond this surface preference and identify the next-best alternative, the option set is \emph{perturbed}. We remove the initial (stage 1) selection $o^*_1$ from $O$ and replace it with a neutral placeholder, \emph{``\textbf{none of the options}''}, yielding the modified option set
\[
O' = (O \setminus \{o^*_1\}) \cup \{\text{none of the options}\}.
\]
The model is then queried again on this modified MCQ, resulting in the second (stage 2) selection $o^*_2 \in O'$. If $o^*_1$ was a robustly correct answer, the model is expected to select its placeholder, \emph{``none of the options''}. Otherwise, the model will select another plausible alternative. The resulting pair of answers, $(o^*_1, o^*_2)$, reveals a structured preference ordering.

\paragraph{Stage 3: Confined Final Inference}
The framework then constructs a simplified MCQ consisting only the two most plausible model-selected candidates.
The final inference is performed over this minimal set, reducing the model's preference instability.

\subsection{Early Stopping Mechanism}
\label{sec:early-stoping}

Our framework incorporates an early stopping mechanism that terminates the process immediately if the most plausible option selected during \textbf{Stage 1} ($o^*_1$) is confirmed in \textbf{Stage 2} (i.e., $o^*_2 =$ \emph{``none of the options''}).
An example of early stopping is presented in \cref{sec:examples-iot}.
\section{Experiments}
Our evaluation focuses on three primary objectives: determining whether IoT delivers consistent accuracy improvements across diverse domains and model scales, comparing its computational efficiency and robustness against existing test-time methods, and diagnosing the specific types of reasoning failures it can (or cannot) mitigate.

\subsection{Setup}

\paragraph{Tasks and Datasets}
We conducted experiments on various MCQ benchmarks. These datasets span diverse domains and difficulty levels, allowing us to validate IoT's generalization across a variety of tasks. The datasets also have different numbers of options, and we can see how performance gain varies on this dimension.

\begin{itemize}[leftmargin=*]
    \item \textbf{Commonsense Reasoning}
    CommonsenseQA~\cite[CSQA,][]{talmor2019commonsenseqaquestionansweringchallenge}. A dataset on commonsense reasoning, with questions in a 5-option MCQs format. 
    OpenbookQA~\cite[OBQA,][]{mihaylov2018suitarmorconductelectricity}. This dataset mimics open-book exams by pairing elementary science questions with a set of core facts. Answering requires multi-step reasoning that combines these provided facts with broad commonsense knowledge.
    SocialIQA~\cite[SIQA,][]{sap2019socialiqacommonsensereasoningsocial}. This dataset is designed to test social commonsense reasoning about people's actions, motivations, and emotional reactions in everyday scenarios. 
    \item \textbf{Educational Benchmark}
    %
    %
    Massive Multi-Task Language Understanding~\cite[MMLU,][]{hendrycks2021measuringmassivemultitasklanguage}, focuses on professional domains such as law, medicine, engineering, and finance. 
    AI2 Reasoning Challenge~\cite[ARC,][]{clark2018thinksolvedquestionanswering}. Derived from U.S. elementary and middle school science exams, it is primarily used to assess the model’s performance in scientific knowledge and complex reasoning tasks.
    \item \textbf{Math Reasoning}
    Algebra Question Answering with Rationales~\cite[AQuA,][]{ling2017programinductionrationalegeneration}. This dataset contains algebra questions, each accompanied by detailed solution steps. It aims to evaluate the model’s problem-solving and logical reasoning abilities in mathematical tasks. 
    GSM8K-MC. This dataset is an MCQ adaptation of the original GSM8K~\cite{cobbe2021trainingverifierssolvemath} dataset, containing elementary-level mathematics problems designed to assess the model’s mathematical reasoning capabilities.
\end{itemize}

\paragraph{Baselines}
We compared the performance gains of the IoT framework against the following baselines. \cref{sec:matched-budget} presents additional results with other baselines with comparable computational cost to IoT.
%

\begin{itemize}[leftmargin=*]
    \item \textbf{Chain-of-Thoughts}~\cite[CoT,][]{wei2023chainofthoughtpromptingelicitsreasoning} is a technique where a language model is prompted to generate a series of short sentences that mimic the reasoning process a person might employ in solving a task. 
    %

    \item \textbf{Self-Consistency}~\cite[SC,][]{wang2023selfconsistencyimproveschainthought} replaces the naive greedy decoding used in chain-of-thought prompting. It first samples a diverse set of reasoning paths instead of only taking the greedy one, and then selects the most consistent answer by marginalizing out the sampled reasoning paths. 
    %

    \item \textbf{Exclusion of Thoughts}~\cite[EoT,][]{fu-etal-2025-exclusion} uses a scoring approach to  eliminate the least important options one at a time until the likelihood of the top option is higher than the other options with respect to a heuristic threshold.
\end{itemize}

While IoT shares a superficial similarity with elimination-based techniques, \emph{e.g.} EoT, it is fundamentally distinguished by several key conceptual and procedural differences:

\paragraph{Focus on Preference Stability vs. Information Reduction} Methods like EoT typically operate on the premise of reducing cognitive load by iteratively removing the least likely distractors based on heuristic scoring. In contrast, IoT probes the model’s internal ranking through the counterfactual removal of its top-rated hypothesis. By isolating the top two candidates between which a model is most likely to oscillate, IoT treats "preference instability" as the primary failure mode rather than a lack of raw reasoning steps.

\paragraph{Computational Efficiency} IoT is designed to be more lightweight than iterative elimination or complex search-based strategies. This efficiency is driven by a dynamic early-stopping mechanism. If the model selects the "none of the options" placeholder in Stage 2, it confirms its initial preference is robustly superior to all alternatives, allowing the process to terminate and bypass Stage 3. Our analysis shows IoT generates approximately 2.2x the tokens of standard CoT, whereas SC requires roughly 4.6x.

\paragraph{Diagnostic Utility} Unlike prior methods, the three-stage structure of IoT enables a granular transition-based analysis. By categorizing model behavior into patterns such as TTT, FTT, or TFF, we can identify whether errors arise from model failure or external factors like genuine question ambiguity and "noisy" benchmark labels. This allows IoT to function as a diagnostic tool for assessing dataset quality in addition to improving performance.

\subsection{Main Results}

As demonstrated in \Cref{tab:main}, IoT achieves consistent accuracy improvements across most benchmarks, attaining the highest average accuracy in each category compared to existing baselines. For instance, when applied to the \texttt{Olmo-2-7B} model, IoT secures the highest overall average accuracy at 72.91\%, surpassing its nearest competitor by a margin of approximately 1.64\%.
The results also reveal specific domain nuances where IoT’s relative gains vary. IoT appears to be less effective when applied to the \texttt{SIQA} dataset. This is due to \texttt{SIQA}'s 3-option format. When the option set is smaller and therefore, distractor competition is low, models encounter less fluctuation, and improvements naturally shrink, as fewer competing hypotheses reduce preference instability.

\begin{figure}[t]
    \centering
    \includegraphics[width=\linewidth]{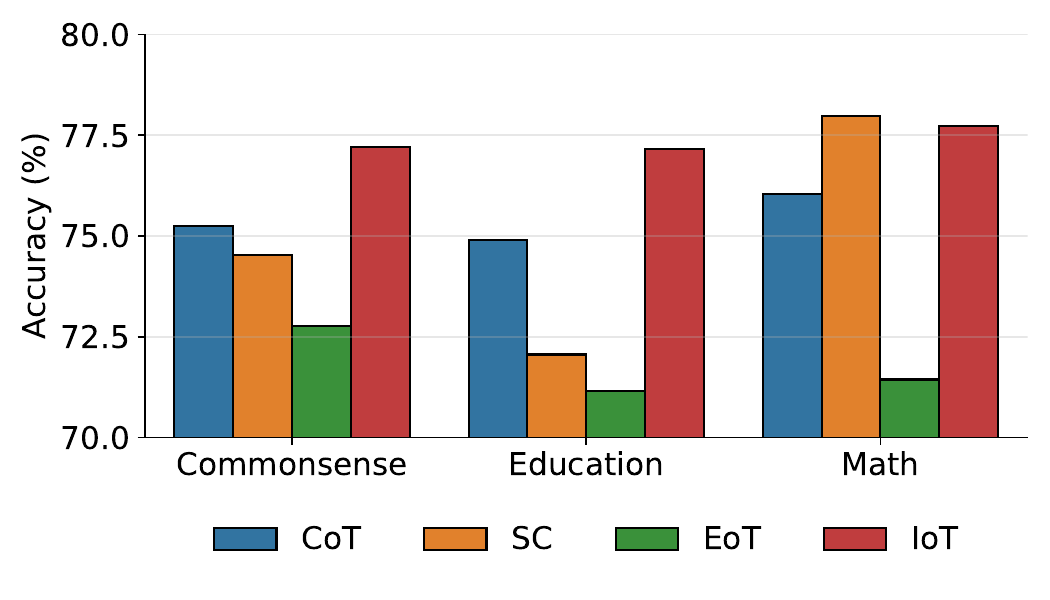}
    \caption{Results of \texttt{Olmo-2-7B} categorized by benchmarks type. While the margin in Commonsense or Education is more significant, IoT achieves performance comparable to self-consistency (SC) with much lower computational cost.}
    \label{fig:categories}
\end{figure}

While IoT yields performance improvements over standard CoT on Math benchmarks, it is less effective than SC. This suggests that for mathematical problems models maybe reasoning without considering the provided options.
In such cases, SC facilitates diverse reasoning pathways through temperature-based sampling, whereas IoT (and other baselines) utilizes a greedy approach that doesn't promote such diversity. All that said, IoT remains the most robust overall method, establishing the highest average performance across all four models.
\Cref{fig:categories} illustrates the performance comparison of IoT with existing baselines across aggregated domain performance.



%
\begin{table*}[t]
\centering
\small
\renewcommand{\arraystretch}{1.25} 

\resizebox{\textwidth}{!}{%
\begin{tabular}{cc ccc cc cc c}
\toprule
\multirow{2}{*}{\textbf{Model}} & \multirow{2}{*}{\textbf{Method}} & \multicolumn{3}{c}{\textbf{Commonsense}} & \multicolumn{2}{c}{\textbf{Education}} & \multicolumn{2}{c}{\textbf{Math}} & \multirow{2}{*}{\textbf{Avg.}}\\ 
\cmidrule(lr){3-5} \cmidrule(lr){6-7} \cmidrule(lr){8-9}
& & \textbf{SIQA (3)} & \textbf{OBQA (4)} & \textbf{CSQA (5)} & \textbf{ARC (4)} & \textbf{MMLU (4)} & \textbf{GSM8K-MC (4)} & \textbf{AQUA (5)} &  \\ \midrule

& \textbf{CoT} & 70.42 & \underline{80.80} & 74.53 & \underline{84.21} & \underline{65.62} & 89.08 & 62.99 & \underline{75.38} \\
& \textbf{SC} & \textbf{71.36} & 75.80 & \underline{76.42} & 79.22 & 64.91 & \underline{89.39} & \textbf{66.54} & 74.81 \\
\multirow{-3}{*}{\rotatebox{90}{\textbf{\scriptsize Olmo-2-7B}}} & \textbf{EoT} & 69.91 & 74.60 & 73.79 & 79.69 & 62.62 & 87.06 & 55.82 & 71.93 \\ \rowcolor[HTML]{EFEFEF}
& \textbf{IoT} & \underline{70.88} & \textbf{84.20} & \textbf{76.58} & \textbf{87.54} & \textbf{66.78} & \textbf{91.66} & \underline{63.78} & \textbf{77.35} \\ \midrule

& \textbf{CoT} & 71.60 & \underline{85.40} & 79.69 & 85.32 & \underline{70.59} & 91.36 & 67.32 & 78.75 \\
& \textbf{SC} & \underline{72.93} & 82.00 & 80.18 & \underline{86.01} & 69.86 & \textbf{95.68} & \textbf{73.62} & \underline{80.04} \\
\multirow{-3}{*}{\rotatebox{90}{\textbf{\scriptsize Olmo-2-13B}}} & \textbf{EoT} & 72.57 & 83.40 & 73.38 & 85.58 & 69.62 & 90.14 & 58.23 & 76.13 \\ \rowcolor[HTML]{EFEFEF}
& \textbf{IoT} & \textbf{73.64} & \textbf{87.60} & \underline{80.01} & \textbf{88.65} & \textbf{71.34} & \underline{93.10} & \underline{70.47} & \textbf{80.69} \\ \midrule

& \textbf{CoT} & 69.75 & 74.60 & 72.89 & 81.06 & 65.56 & 87.41 & 54.72 & 72.28 \\
& \textbf{SC} & \textbf{71.14} & \underline{77.80} & \underline{74.77} & \underline{82.57} & \textbf{69.00} & \textbf{91.89} & \underline{57.09} & \underline{74.89} \\
\multirow{-3}{*}{\rotatebox{90}{\textbf{\scriptsize Llama-3.3-8B}}} & \textbf{EoT} & \underline{70.57} & 75.80 & 74.44 & 81.99 & \underline{66.56} & 84.31 & 53.01 & 72.38 \\ \rowcolor[HTML]{EFEFEF}
& \textbf{IoT} & 69.70 & \textbf{78.40} & \textbf{75.84} & \textbf{83.79} & 66.31 & \underline{90.67} & \textbf{61.81} & \textbf{75.22} \\ \midrule

& \textbf{CoT} & \underline{79.52} & \underline{88.80} & \underline{83.21} & \underline{94.54} & \underline{81.31} & \underline{94.74} & \underline{74.37} & \underline{85.23} \\ \rowcolor[HTML]{EFEFEF}
\multirow{-2}{*}{\rotatebox{90}{\textbf{\scriptsize \shortstack{GPT-4o\\mini}}}} & \textbf{IoT} & \textbf{80.04} & \textbf{93.40} & \textbf{84.54} & \textbf{94.62} & \textbf{82.06} & \textbf{95.00} & \textbf{79.13} & \textbf{86.21} \\ \bottomrule

\end{tabular}
}
\caption{Accuracy of multiple LLMs across various benchmarks using CoT, Self-Consistency (SC), EoT, detecting plausible answers, and IoT. 
The second best performance is \underline{underlined}. \textbf{Note:} EoT cannot be applied directly to closed-source models (GPT), and GPT-4o SC is excluded due to computational costs.}
\label{tab:main}
\end{table*}

\subsection{Transition Analysis}
\label{sec:transition}

\begin{figure}
    \centering
    \includegraphics[width=0.9\linewidth]{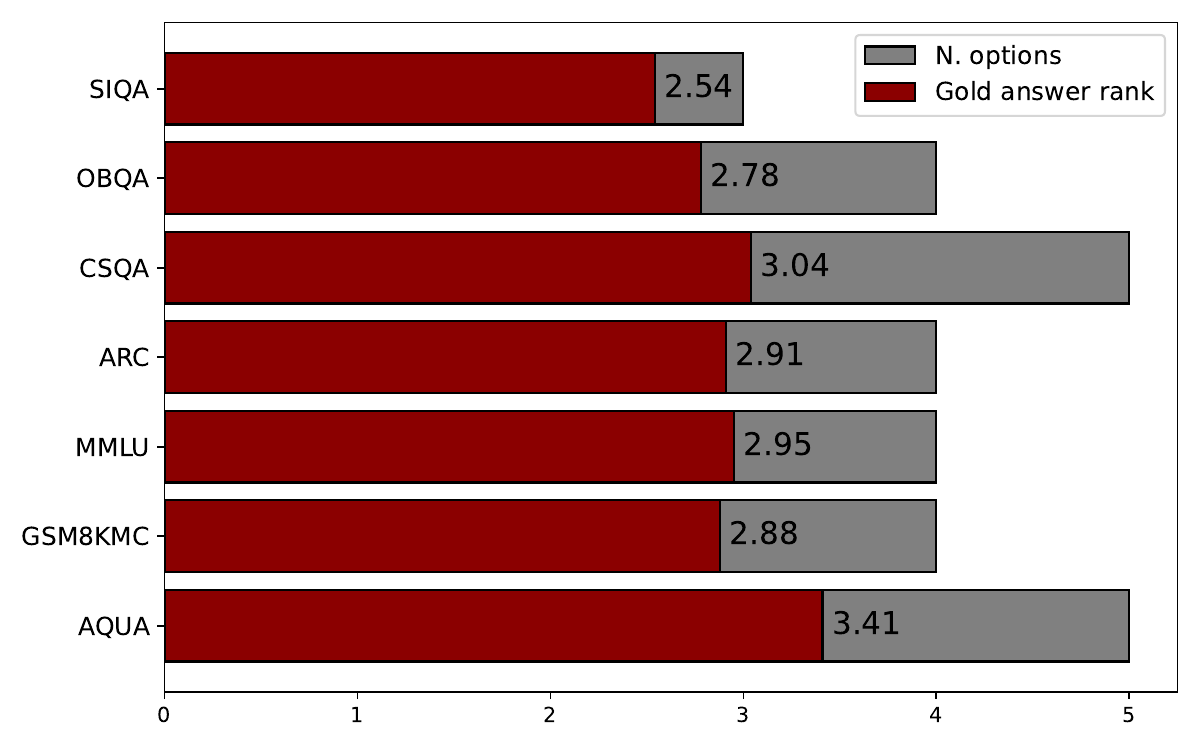}
    \caption{Gold answer average ranking for FFF cases.}
    \label{fig:fff-rank}
\end{figure}

Beyond aggregate accuracy, IoT enables a fine-grained analysis of how model predictions evolve under option perturbation, revealing distinct transition patterns that characterize different reasoning failure modes.
During each stage, the model output can be either True (\emph{i.e.} it matches the ground truth) or False (F). Therefore, from stage 1 to stage 3, the model answer can flip multiple times, and it can be categorized into the following groups:

\begin{itemize}[leftmargin=*]
    \item \textbf{TTT:} When the model detects and retains its correct answer during all stages, it indicates that the model has high confidence in the correct answer. 
    \item \textbf{FFF:} In some cases, the model is not able to detect the correct answer in stages 1 and 2, and therefore stage 3.. \Cref{fig:fff-rank} shows that in FFF cases the gold answer does not appear within the model's top prediction, meaning that it has weak knowledge about the gold answer. 
    \item \textbf{TFT, FTT:} In these cases, the correct answer appears in either stage 1 or 2, and the model successfully chose the correct answer in stage 3.
    \item \textbf{TFF, FTF:} In these cases, the correct answers appear in either stage 1 or 2. However, IoT fails to recover the correct answer in stage 3.
\end{itemize}

\begin{figure}[t]
    \centering
    \includegraphics[width=0.6\columnwidth]{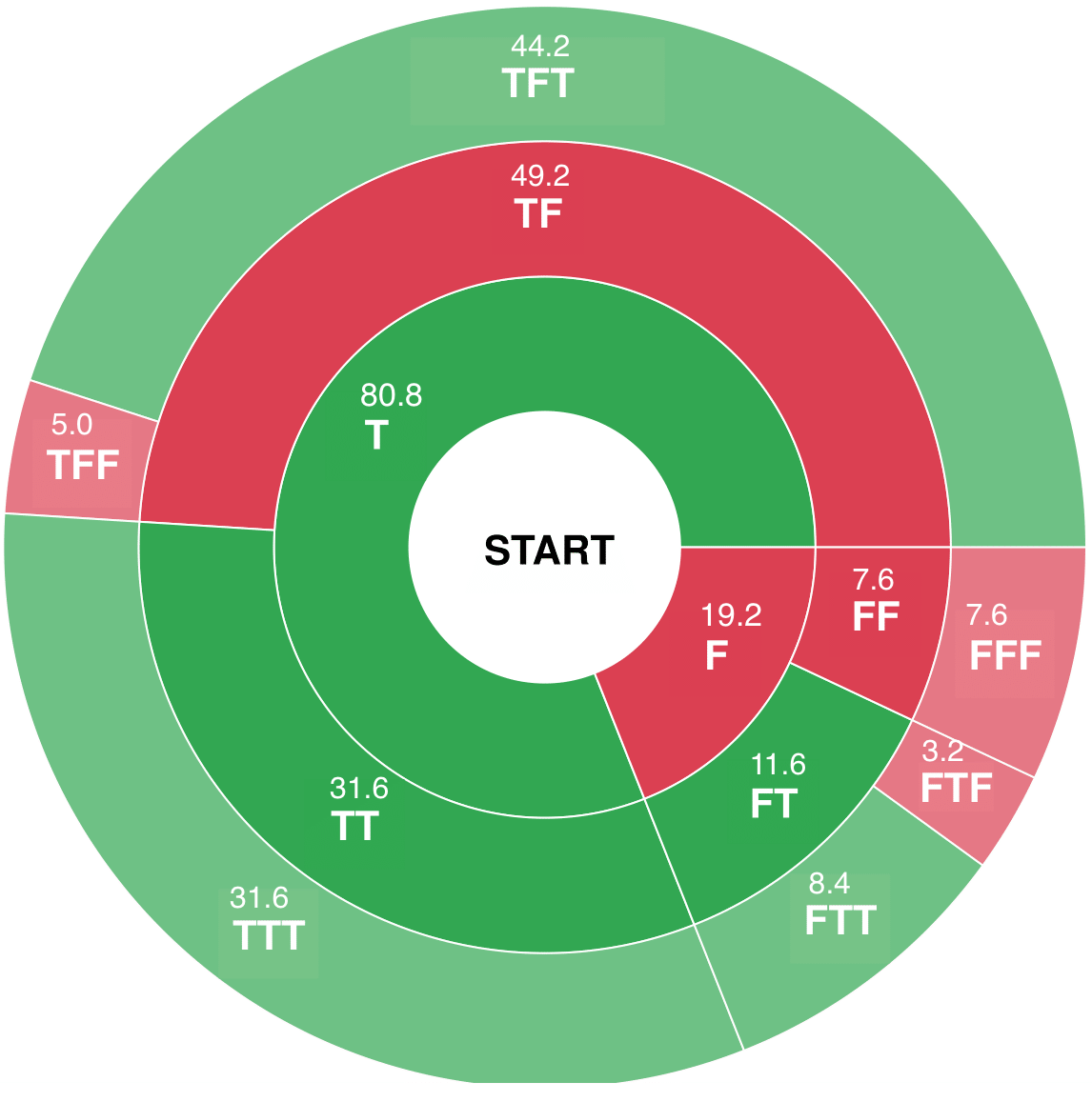}
    \caption{Answer transition process of \texttt{Olmo-2-7B} over OBQA dataset. The final improvement is the difference between FTT and TFF.}
    \label{fig:progress}
\end{figure}

As shown in \cref{fig:progress}, the performance improvement of IoT (outermost layer) w.r.t. CoT (innermost layer) is due to transforming some of the initially incorrect answers into correct answers (FTT) minus a small portion of correct answers turning into false (TFF). 

We next look into the TFF and FTF cases. These instances are interesting because the model identifies the correct answer at one stage but selects the incorrect answer in the end. We suspect this may be due to noisy ground truth: perhaps they are both plausible answers or the ground truth answer is incorrect.
We show an example of each case: 

\begin{tcolorbox}[
        colback=white!25!white,
        colframe=gray!100!gray,
        colbacktitle=red!75!black, 
        title=\small FTF - Multiple groundtruth,
        fonttitle=\bfseries, 
        width=\columnwidth
        ]
\small
\label{}
She was always helping at the senior center, it brought her what? 
\textbf{(A) happiness} (B) satisfaction
\end{tcolorbox}

\begin{tcolorbox}[
        colback=white!25!white,
        colframe=gray!100!gray,
        colbacktitle=red!75!black, 
        title=\small FTF - Wrong groundtruth,
        fonttitle=\bfseries, 
        width=\columnwidth
        ]
\small
\label{}
What could committing murder prevent someone from doing? \textbf{(A) find god} (B) go to jail
\end{tcolorbox}

\begin{figure}[h]
    \centering
    \includegraphics[width=0.9\linewidth]{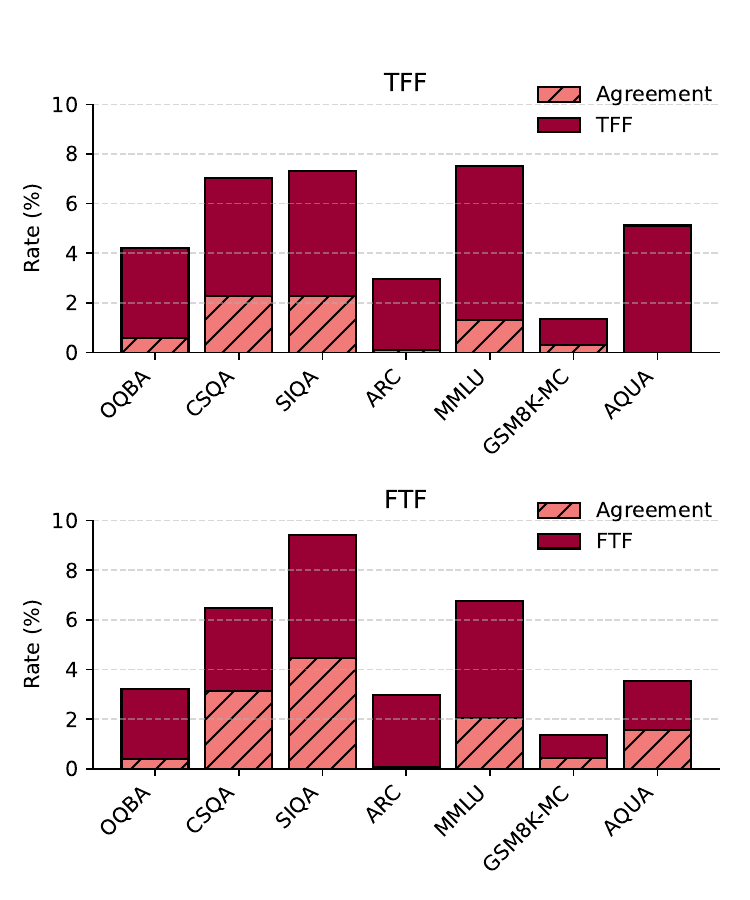}
    \caption{The portion of noisy samples agreed across different benchmarks in TFF and FTF agreed by external judges. The Agreement rate shown in the plot is the average rate over \texttt{Gemini-3-pro}, and \texttt{GPT-5} agreement. }
    \label{fig:transition}
\end{figure}

To assess these cases (TFF and FTF) quantitatively, we use \texttt{Gemini-3-pro} and \texttt{GPT-5} as LLM judges. Specifically, we use stage 3 questions with two model-selected options and ask the judges to select the correct answer. The rate at which the judges' selected answer aligns with the IoT final output is recorded as the agreement rate. 
%
As shown in \cref{fig:transition}, the judges' evaluation suggests that perhaps a substantial portion of these ``error cases'' are perhaps not error after all (notwithstanding the variance across datasets) and that the performance we saw in \Cref{tab:main} is a lower bound of its true performance.

%
%

\subsection{Matched-budget Baselines}
\label{sec:matched-budget}
The simplicity of IoT is a major strength. To better contextualise this property, we compare IoT with two simple alternatives that have comparable computational cost to IoT: (1) top-1 + one random selection, which retains the highest-scoring option and one randomly selected alternative, thereby eliminating the computational cost of Stage 2; and (2) logit-based filtering, a simple heuristic that selects the two options with the highest likelihood scores. 

Since Stage 3 is the same across all baselines, we only compare the top-2 accuracy. As shown in \cref{tab:matched-budget}, IoT significantly improves top-2 accuracy compared to random second-option selection. Importantly, this improvement translates into higher final-stage accuracy, indicating that the gains are not merely due to additional opportunities but arise from structured counterfactual perturbation. 

We also further evaluated SC with 3 samples (closer to IoT in token cost) for \texttt{Olmo-2-7b}. As can be seen in \cref{tab:sc}, IoT remains competitive or superior on most non-math tasks at similar budgets.

\begin{table}[t]
\centering
\small 
\resizebox{\columnwidth}{!}{%
\begin{tabular}{l ccc} 
\toprule
\textbf{Dataset}& \textbf{top-1+random}& \textbf{logit-based} & \textbf{IoT}\\  \midrule
SIQA      & 75.79 & \textbf{90.78} & \underline{88.02} \\
OBQA      & 83.00 & 86.80 & \textbf{92.40} \\
CSQA      & 79.28 & 87.14 & \textbf{89.19} \\
ARC       & 86.94 & 89.85 & \textbf{93.86} \\ 
GSM8K-MC  & 91.89 & 58.68 & \textbf{94.54} \\
AQUA      & \textbf{74.02} & 42.23 & \underline{73.23} \\ \bottomrule
\end{tabular}}
\caption{Top-2 accuracy of match-budget baselines compared to IoT.}
\label{tab:matched-budget}
\end{table}

\begin{table}[t]
\centering
\small 
\resizebox{\columnwidth}{!}{%
\begin{tabular}{l cccc} 
\toprule
\textbf{Dataset}& \textbf{CoT}& \textbf{SC (3 samples)}& \textbf{SC (5 samples)} & \textbf{IoT}\\  \midrule
SIQA      & 70.42 & 68.73 & \textbf{71.36} & 70.88 \\
OBQA      & 80.80 & 80.80 & 75.80 & \textbf{84.20} \\
CSQA      & 74.53 & 74.12 & 76.42 & \textbf{76.58} \\
ARC       & 84.21 & 84.38 & 79.22 & \textbf{87.54} \\ 
MMLU      & 65.62 & N/A   & 64.91 & \textbf{66.78} \\ 
GSM8K-MC  & 89.08 & \textbf{93.32} & 89.39 & 91.66 \\
AQUA      & 62.99 & 65.35 & \textbf{66.54} & 63.78 \\ \bottomrule
\end{tabular}}
\caption{Comparison of self-consistency with different numbers of samples to IoT.}
\label{tab:sc}
\end{table}

\subsection{More Chances, Less Performance}

To determine the optimal number of chances that maximizes the performance, we extend our framework to allow the model a potential opportunity at option selection. This modified process filters the candidate options down to a maximum of \emph{three} distinct choices (or fewer, if the model repeats previous selections). The final question is then constructed using these filtered options.

\begin{table}[t]
\centering
\resizebox{\columnwidth}{!}{%
\begin{tabular}{@{}l c c c@{}} 
\toprule
\multirow{2}{*}{\textbf{Dataset}} & \multirow{2}{*}{\textbf{CoT}} & \multicolumn{2}{c}{\textbf{Inclusion-of-Thoughts}} \\ 
\cmidrule(lr){3-4} 
 & & \textbf{w/ 2 Chances} & \textbf{w/ 3 Chances} \\ \midrule
SIQA       & 70.09 & \textbf{70.29} & N/A   \\
OBQA       & 77.70 & \textbf{81.30} & 80.70 \\
CSQA       & 73.34 & \textbf{76.41} & 75.67 \\
ARC        & 82.64 & \textbf{85.67} & 84.81 \\
MMLU       & 65.59 & 66.54 & \textbf{66.92} \\
GSM8K-MC   & 88.25 & \textbf{91.17} & 89.50 \\
AQUA       & 58.86 & \textbf{62.80} & 60.04 \\ \bottomrule
\end{tabular}
}
\caption{Comparison of IoT performance given a second and third chance. The results are average over \texttt{Olmo-2-7B} and \texttt{Llama-3.3-8B} performance.}
\label{tab:third-chance}
\end{table}

As demonstrated in \cref{tab:third-chance}, model performance peaks when the selection process is limited to a second chance, and degrades when a third chance is introduced. This diminishing return can be attributed to two main factors:
(1) If the correct answer is in the top two predictions, then it means it has a low model probability. And so the likelihood of the model recovering its performance in the final stage (when given 3 options) is low.
%
(2) Increasing the number of option inherently increases the number of distractors in the final stage, and this can negatively impact model decision making (in other words, there is an opportunity cost for adding every additional candidate).


\begin{figure*}[t]
    \centering
    \includegraphics[width=\linewidth]{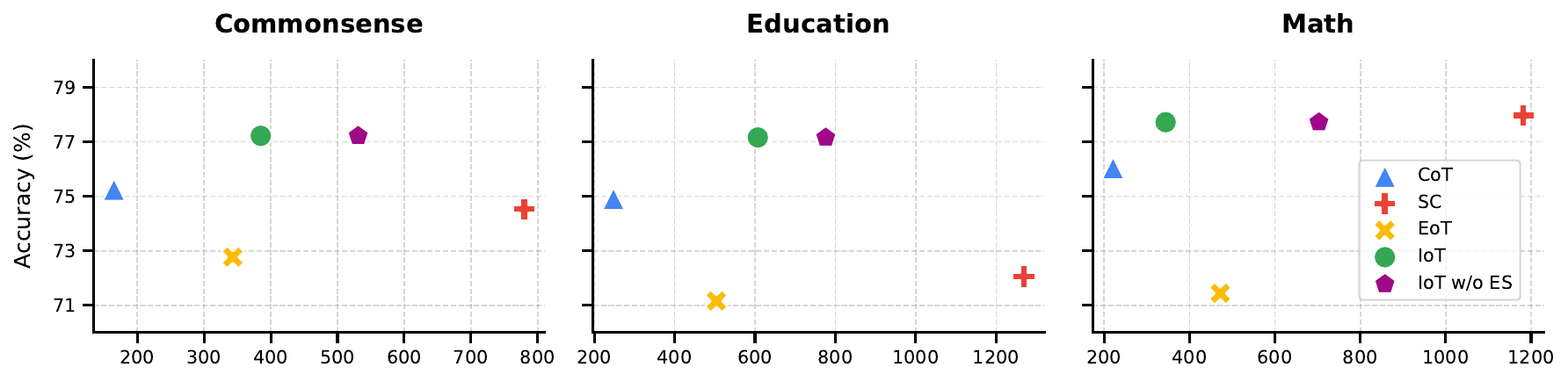}
    \caption{Cost-Benefit. Vertical axes illustrate the accuracy of each method, while the horizontal axes show the average number of tokens per sample that each method costs. ``IoT w/o ES'' means IoT without early stopping. The detailed number of tokens generated by each dataset is given in \cref{sec:overhead}.}
    \label{fig:cost-benefit}
\end{figure*}

\subsection{Robustness}
\citet{pezeshkpour2023largelanguagemodelssensitivity} reveal that LLMs exhibit a ``sensitivity gap'', where model predictions shift based on the positional order of choices in multiple-choice benchmarks. 
To ensure IoT provides a reliable measure of task performance, we conduct a robustness analysis by systematically shuffling the order of options 5 times and calculating the variance in performance across permutations. We also compute the selection entropy, where we measure how consistently the model stays locked onto the correct semantic choice versus getting pulled away by positional bias.
We aim to quantify the extent to which our framework mitigates these ordering effects and maintains a consistent, label-invariant representation of the underlying knowledge. \Cref{tab:robustness} results indicate that IoT not only improves average accuracy but also reduces sensitivity to option ordering, suggesting a more stable underlying preference structure.
%

\begin{table}[t]
\centering
\resizebox{\columnwidth}{!}{%
\begin{tabular}{l ccc ccc} 
\toprule
\multirow{2}{*}{\textbf{Dataset}} & \multicolumn{3}{c}{\textbf{CoT}} & \multicolumn{3}{c}{\textbf{IoT}} \\ 
\cmidrule(lr){2-4} \cmidrule(lr){5-7} 
 & \textbf{Mean} & \textbf{Std} & \textbf{Entropy} & \textbf{Mean} & \textbf{Std} & \textbf{Entropy} \\ \midrule

SIQA     & 69.63 & \textbf{0.49} & 0.76 & \textbf{71.89} & 0.61 & \textbf{0.43}\\
OBQA     & 80.64 & 0.84          & 0.79 & \textbf{84.28}   & \textbf{0.73} & \textbf{0.24}\\
CSQA     & 72.64 & 1.13          & 0.72 & \textbf{76.89}   & \textbf{0.52} & \textbf{0.19}\\
ARC      & 83.43 & \textbf{0.50} & 0.78 & \textbf{86.82}   & 0.53          & \textbf{0.26}\\
MMLU     & 64.58 & 0.64          & 0.72 & \textbf{66.84}   & \textbf{0.29} & \textbf{0.27}\\
GSM8K-MC & 90.19 & 0.74          & 0.82 & \textbf{92.70}   & \textbf{0.64} & \textbf{0.22}\\
AQUA     & 58.11 & 3.53          & 0.64 & \textbf{63.93}   & \textbf{2.02} & \textbf{0.20}\\ \bottomrule

\end{tabular}
}
\caption{Robustness of IoT compared to CoT w.r.t. shuffling options. Results are obtained from \texttt{Olmo-2-7B}. The best performance is \textbf{bold}.}
\label{tab:robustness}
\end{table}






\subsection{Computational Overhead}
Despite the iterative nature of the IoT, it introduces only modest computational overhead in practice.
This is because many instances do not go through the three stages  if the same option is selected in stage 1 and 2 (early stopping; \cref{sec:early-stoping}).
%
\cref{fig:cost-benefit} presents the cost-benefit trade-off where we plot task performance vs.\ number of tokens generated by IoT and baseline methods on \texttt{Olmo-2-7B}. The results show that IoT produces a good trade-off in terms of efficiency vs. performance.
\Cref{sec:overhead} provides a more detailed breakdown of the token cost.

\subsection{IoT Combined with Existing Baselines}
Given that self-consistency (SC) is the best-performing baseline, we explored the complementary strengths of IoT. In \cref{tab:complement}, we compare the performance of CoT, SC, IoT, and the combination of IoT+SC. For IoT+SC, we follow the same procedure outlined in \cref{sec:iot} – the only difference is that we apply SC to obtain the answer at each stage. We find that while both SC and IoT yield individual improvements over CoT, their combination yields slightly greater enhancements compared to either IoT or SC alone, indicating that IoT effectively complements existing methods such as SC. However, IoT+SC imposes high computational costs for only marginal performance gains.

\begin{table}[t]
\centering
\small
\begin{tabular*}{\columnwidth}{@{\extracolsep{\fill}} l rrrr} 
\toprule
\textbf{Dataset} & \textbf{CoT} & \textbf{SC} & \textbf{IoT} & \textbf{IoT+SC} \\ \midrule
SIQA      & 70.42 & 71.36 & 70.88 & 66.63 \\
OBQA      & 80.80 & 75.80 & 84.20 & 84.40 \\
CSQA      & 74.53 & 76.42 & 76.58 & 71.82 \\
ARC       & 84.21 & 79.22 & 87.54 & 83.53 \\ 
GSM8K-MC  & 89.39 & 89.39 & 91.66 & 93.02 \\
AQUA      & 62.99 & 66.54 & 63.78 & 64.96 \\ \bottomrule
\end{tabular*}

\caption{IoT complements existing methods such as SC. While SC shows improvements by sampling more forward reasoning, integrating IoT further improves the performance.}
\label{tab:complement}
\end{table}

\section{Conclusion}
We introduced the IoT framework, a novel prompting strategy that systematically eliminates incorrect options to enhance LLM performance on MCQ tasks. Inspired by human reasoning strategies, IoT reduces distractors by redirecting the model’s attention to critical options, allowing it to focus on the relevant choices. Our experiments across a variety of MCQ datasets demonstrate that IoT improves reasoning accuracy. We believe that viewing MC reasoning through the lens of preference stability opens new directions for understanding and improving LLM reasoning behavior under controlled perturbations.

\section*{Limitations}
IoT is designed specifically for MCQ question answering and may not generalize to open-ended generation tasks.
Its effectiveness diminishes when the option set is very small, as preference instability is less pronounced in such settings.
%
%
Finally, while IoT improves stability at inference time, it does not address errors caused by missing knowledge or fundamentally incorrect reasoning.


\section*{Ethics Statement}
All experiments were conducted using publicly available datasets. 
%
%
The LLMs in the study were used in accordance with their terms of service. 
AI assistants were used to help writing and refining the paper.

\bibliography{custom}

@misc{talmor2019commonsenseqaquestionansweringchallenge,
      title={CommonsenseQA: A Question Answering Challenge Targeting Commonsense Knowledge}, 
      author={Alon Talmor and Jonathan Herzig and Nicholas Lourie and Jonathan Berant},
      year={2019},
      eprint={1811.00937},
      archivePrefix={arXiv},
      primaryClass={cs.CL},
      url={https://arxiv.org/abs/1811.00937}, 
}

@misc{mihaylov2018suitarmorconductelectricity,
      title={Can a Suit of Armor Conduct Electricity? A New Dataset for Open Book Question Answering}, 
      author={Todor Mihaylov and Peter Clark and Tushar Khot and Ashish Sabharwal},
      year={2018},
      eprint={1809.02789},
      archivePrefix={arXiv},
      primaryClass={cs.CL},
      url={https://arxiv.org/abs/1809.02789}, 
}

@misc{sap2019socialiqacommonsensereasoningsocial,
      title={SocialIQA: Commonsense Reasoning about Social Interactions}, 
      author={Maarten Sap and Hannah Rashkin and Derek Chen and Ronan LeBras and Yejin Choi},
      year={2019},
      eprint={1904.09728},
      archivePrefix={arXiv},
      primaryClass={cs.CL},
      url={https://arxiv.org/abs/1904.09728}, 
}

@misc{clark2018thinksolvedquestionanswering,
      title={Think you have Solved Question Answering? Try ARC, the AI2 Reasoning Challenge}, 
      author={Peter Clark and Isaac Cowhey and Oren Etzioni and Tushar Khot and Ashish Sabharwal and Carissa Schoenick and Oyvind Tafjord},
      year={2018},
      eprint={1803.05457},
      archivePrefix={arXiv},
      primaryClass={cs.AI},
      url={https://arxiv.org/abs/1803.05457}, 
}

@misc{ling2017programinductionrationalegeneration,
      title={Program Induction by Rationale Generation : Learning to Solve and Explain Algebraic Word Problems}, 
      author={Wang Ling and Dani Yogatama and Chris Dyer and Phil Blunsom},
      year={2017},
      eprint={1705.04146},
      archivePrefix={arXiv},
      primaryClass={cs.AI},
      url={https://arxiv.org/abs/1705.04146}, 
}

@misc{cobbe2021trainingverifierssolvemath,
      title={Training Verifiers to Solve Math Word Problems}, 
      author={Karl Cobbe and Vineet Kosaraju and Mohammad Bavarian and Mark Chen and Heewoo Jun and Lukasz Kaiser and Matthias Plappert and Jerry Tworek and Jacob Hilton and Reiichiro Nakano and Christopher Hesse and John Schulman},
      year={2021},
      eprint={2110.14168},
      archivePrefix={arXiv},
      primaryClass={cs.LG},
      url={https://arxiv.org/abs/2110.14168}, 
}

@misc{wei2023chainofthoughtpromptingelicitsreasoning,
      title={Chain-of-Thought Prompting Elicits Reasoning in Large Language Models}, 
      author={Jason Wei and Xuezhi Wang and Dale Schuurmans and Maarten Bosma and Brian Ichter and Fei Xia and Ed Chi and Quoc Le and Denny Zhou},
      year={2023},
      eprint={2201.11903},
      archivePrefix={arXiv},
      primaryClass={cs.CL},
      url={https://arxiv.org/abs/2201.11903}, 
}

@misc{wang2023selfconsistencyimproveschainthought,
      title={Self-Consistency Improves Chain of Thought Reasoning in Language Models}, 
      author={Xuezhi Wang and Jason Wei and Dale Schuurmans and Quoc Le and Ed Chi and Sharan Narang and Aakanksha Chowdhery and Denny Zhou},
      year={2023},
      eprint={2203.11171},
      archivePrefix={arXiv},
      primaryClass={cs.CL},
      url={https://arxiv.org/abs/2203.11171}, 
}

@misc{ma2023poeprocesseliminationmultiple,
      title={POE: Process of Elimination for Multiple Choice Reasoning}, 
      author={Chenkai Ma and Xinya Du},
      year={2023},
      eprint={2310.15575},
      archivePrefix={arXiv},
      primaryClass={cs.CL},
      url={https://arxiv.org/abs/2310.15575}, 
}

@misc{andor2019givingbertcalculatorfinding,
      title={Giving BERT a Calculator: Finding Operations and Arguments with Reading Comprehension}, 
      author={Daniel Andor and Luheng He and Kenton Lee and Emily Pitler},
      year={2019},
      eprint={1909.00109},
      archivePrefix={arXiv},
      primaryClass={cs.CL},
      url={https://arxiv.org/abs/1909.00109}, 
}

@misc{ran2019numnetmachinereadingcomprehension,
      title={NumNet: Machine Reading Comprehension with Numerical Reasoning}, 
      author={Qiu Ran and Yankai Lin and Peng Li and Jie Zhou and Zhiyuan Liu},
      year={2019},
      eprint={1910.06701},
      archivePrefix={arXiv},
      primaryClass={cs.CL},
      url={https://arxiv.org/abs/1910.06701}, 
}

@misc{geva2020injectingnumericalreasoningskills,
      title={Injecting Numerical Reasoning Skills into Language Models}, 
      author={Mor Geva and Ankit Gupta and Jonathan Berant},
      year={2020},
      eprint={2004.04487},
      archivePrefix={arXiv},
      primaryClass={cs.CL},
      url={https://arxiv.org/abs/2004.04487}, 
}

@misc{piękos2021measuringimprovingbertsmathematical,
      title={Measuring and Improving BERT's Mathematical Abilities by Predicting the Order of Reasoning}, 
      author={Piotr Piękos and Henryk Michalewski and Mateusz Malinowski},
      year={2021},
      eprint={2106.03921},
      archivePrefix={arXiv},
      primaryClass={cs.CL},
      url={https://arxiv.org/abs/2106.03921}, 
}

@misc{madaan2023selfrefineiterativerefinementselffeedback,
      title={Self-Refine: Iterative Refinement with Self-Feedback}, 
      author={Aman Madaan and Niket Tandon and Prakhar Gupta and Skyler Hallinan and Luyu Gao and Sarah Wiegreffe and Uri Alon and Nouha Dziri and Shrimai Prabhumoye and Yiming Yang and Shashank Gupta and Bodhisattwa Prasad Majumder and Katherine Hermann and Sean Welleck and Amir Yazdanbakhsh and Peter Clark},
      year={2023},
      eprint={2303.17651},
      archivePrefix={arXiv},
      primaryClass={cs.CL},
      url={https://arxiv.org/abs/2303.17651}, 
}

@misc{hendrycks2021measuringmassivemultitasklanguage,
      title={Measuring Massive Multitask Language Understanding}, 
      author={Dan Hendrycks and Collin Burns and Steven Basart and Andy Zou and Mantas Mazeika and Dawn Song and Jacob Steinhardt},
      year={2021},
      eprint={2009.03300},
      archivePrefix={arXiv},
      primaryClass={cs.CY},
      url={https://arxiv.org/abs/2009.03300}, 
}

@article{llama3modelcard,
    title={Llama 3 Model Card},
    author={AI@Meta},
    year={2024},
    url = {https://github.com/meta-llama/llama3/blob/main/MODEL_CARD.md}
}

@article{olmo20242olmo2furious,
      title={2 OLMo 2 Furious}, 
      author={Team OLMo and Pete Walsh and Luca Soldaini and Dirk Groeneveld and Kyle Lo and Shane Arora and Akshita Bhagia and Yuling Gu and Shengyi Huang and Matt Jordan and Nathan Lambert and Dustin Schwenk and Oyvind Tafjord and Taira Anderson and David Atkinson and Faeze Brahman and Christopher Clark and Pradeep Dasigi and Nouha Dziri and Michal Guerquin and Hamish Ivison and Pang Wei Koh and Jiacheng Liu and Saumya Malik and William Merrill and Lester James V. Miranda and Jacob Morrison and Tyler Murray and Crystal Nam and Valentina Pyatkin and Aman Rangapur and Michael Schmitz and Sam Skjonsberg and David Wadden and Christopher Wilhelm and Michael Wilson and Luke Zettlemoyer and Ali Farhadi and Noah A. Smith and Hannaneh Hajishirzi},
      year={2024},
      eprint={2501.00656},
      archivePrefix={arXiv},
      primaryClass={cs.CL},
      url={https://arxiv.org/abs/2501.00656}, 
}

@inproceedings{lyu2023faithful,
  title={Faithful chain-of-thought reasoning},
  author={Lyu, Qing and Havaldar, Shreya and Stein, Adam and Zhang, Li and Rao, Delip and Wong, Eric and Apidianaki, Marianna and Callison-Burch, Chris},
  booktitle={The 13th International Joint Conference on Natural Language Processing and the 3rd Conference of the Asia-Pacific Chapter of the Association for Computational Linguistics (IJCNLP-AACL 2023)},
  year={2023}
}

@article{lanham2023measuring,
  title={Measuring faithfulness in chain-of-thought reasoning},
  author={Lanham, Tamera and Chen, Anna and Radhakrishnan, Ansh and Steiner, Benoit and Denison, Carson and Hernandez, Danny and Li, Dustin and Durmus, Esin and Hubinger, Evan and Kernion, Jackson and others},
  journal={arXiv preprint arXiv:2307.13702},
  year={2023}
}

@article{saparov2022language,
  title={Language models are greedy reasoners: A systematic formal analysis of chain-of-thought},
  author={Saparov, Abulhair and He, He},
  journal={arXiv preprint arXiv:2210.01240},
  year={2022}
}

@article{michael2023debate,
  title={Debate helps supervise unreliable experts},
  author={Michael, Julian and Mahdi, Salsabila and Rein, David and Petty, Jackson and Dirani, Julien and Padmakumar, Vishakh and Bowman, Samuel R},
  journal={arXiv preprint arXiv:2311.08702},
  year={2023}
}

@inproceedings{jung2022maieutic,
  title={Maieutic prompting: Logically consistent reasoning with recursive explanations},
  author={Jung, Jaehun and Qin, Lianhui and Welleck, Sean and Brahman, Faeze and Bhagavatula, Chandra and Le Bras, Ronan and Choi, Yejin},
  booktitle={Proceedings of the 2022 conference on empirical methods in natural language processing},
  pages={1266--1279},
  year={2022}
}

@article{yao2023tree,
  title={Tree of thoughts: Deliberate problem solving with large language models},
  author={Yao, Shunyu and Yu, Dian and Zhao, Jeffrey and Shafran, Izhak and Griffiths, Tom and Cao, Yuan and Narasimhan, Karthik},
  journal={Advances in neural information processing systems},
  volume={36},
  pages={11809--11822},
  year={2023}
}

@article{shinn2023reflexion,
  title={Reflexion: Language agents with verbal reinforcement learning},
  author={Shinn, Noah and Cassano, Federico and Gopinath, Ashwin and Narasimhan, Karthik and Yao, Shunyu},
  journal={Advances in Neural Information Processing Systems},
  volume={36},
  pages={8634--8652},
  year={2023}
}

@inproceedings{weng2023large,
  title={Large language models are better reasoners with self-verification},
  author={Weng, Yixuan and Zhu, Minjun and Xia, Fei and Li, Bin and He, Shizhu and Liu, Shengping and Sun, Bin and Liu, Kang and Zhao, Jun},
  booktitle={Findings of the Association for Computational Linguistics: EMNLP 2023},
  pages={2550--2575},
  year={2023}
}

@article{xue2023rcot,
  title={Rcot: Detecting and rectifying factual inconsistency in reasoning by reversing chain-of-thought},
  author={Xue, Tianci and Wang, Ziqi and Wang, Zhenhailong and Han, Chi and Yu, Pengfei and Ji, Heng},
  journal={arXiv preprint arXiv:2305.11499},
  year={2023}
}

@inproceedings{jiang2024forward,
  title={Forward-backward reasoning in large language models for mathematical verification},
  author={Jiang, Weisen and Shi, Han and Yu, Longhui and Liu, Zhengying and Zhang, Yu and Li, Zhenguo and Kwok, James},
  booktitle={Findings of the Association for Computational Linguistics: ACL 2024},
  pages={6647--6661},
  year={2024}
}

@inproceedings{fu-etal-2025-exclusion,
    title = "Exclusion of Thought: Mitigating Cognitive Load in Large Language Models for Enhanced Reasoning in Multiple-Choice Tasks",
    author = "Fu, Qihang  and
      Qin, Yongbin  and
      Huang, Ruizhang  and
      Chen, Yanping  and
      Zhou, Yulin  and
      Long, Lintao",
    editor = "Che, Wanxiang  and
      Nabende, Joyce  and
      Shutova, Ekaterina  and
      Pilehvar, Mohammad Taher",
    booktitle = "Proceedings of the 63rd Annual Meeting of the Association for Computational Linguistics (Volume 1: Long Papers)",
    month = jul,
    year = "2025",
    address = "Vienna, Austria",
    publisher = "Association for Computational Linguistics",
    url = "https://aclanthology.org/2025.acl-long.1051/",
    doi = "10.18653/v1/2025.acl-long.1051",
    pages = "21673--21686",
    ISBN = "979-8-89176-251-0"
}

@misc{pezeshkpour2023largelanguagemodelssensitivity,
      title={Large Language Models Sensitivity to The Order of Options in Multiple-Choice Questions}, 
      author={Pouya Pezeshkpour and Estevam Hruschka},
      year={2023},
      eprint={2308.11483},
      archivePrefix={arXiv},
      primaryClass={cs.CL},
      url={https://arxiv.org/abs/2308.11483}, 
}

@inproceedings{balepur2024s,
  title={It’s not easy being wrong: Large language models struggle with process of elimination reasoning},
  author={Balepur, Nishant and Palta, Shramay and Rudinger, Rachel},
  booktitle={Findings of the Association for Computational Linguistics: ACL 2024},
  pages={10143--10166},
  year={2024}
}

@misc{openai2024gpt4ocard,
      title={GPT-4o System Card}, 
      author={OpenAI and : and Aaron Hurst and Adam Lerer and Adam P. Goucher and Adam Perelman and Aditya Ramesh and Aidan Clark and AJ Ostrow and Akila Welihinda and Alan Hayes and Alec Radford and Aleksander Mądry and Alex Baker-Whitcomb and Alex Beutel and Alex Borzunov and Alex Carney and Alex Chow and Alex Kirillov and Alex Nichol and Alex Paino and Alex Renzin and Alex Tachard Passos and Alexander Kirillov and Alexi Christakis and Alexis Conneau and Ali Kamali and Allan Jabri and Allison Moyer and Allison Tam and Amadou Crookes and Amin Tootoochian and Amin Tootoonchian and Ananya Kumar and Andrea Vallone and Andrej Karpathy and Andrew Braunstein and Andrew Cann and Andrew Codispoti and Andrew Galu and Andrew Kondrich and Andrew Tulloch and Andrey Mishchenko and Angela Baek and Angela Jiang and Antoine Pelisse and Antonia Woodford and Anuj Gosalia and Arka Dhar and Ashley Pantuliano and Avi Nayak and Avital Oliver and Barret Zoph and Behrooz Ghorbani and Ben Leimberger and Ben Rossen and Ben Sokolowsky and Ben Wang and Benjamin Zweig and Beth Hoover and Blake Samic and Bob McGrew and Bobby Spero and Bogo Giertler and Bowen Cheng and Brad Lightcap and Brandon Walkin and Brendan Quinn and Brian Guarraci and Brian Hsu and Bright Kellogg and Brydon Eastman and Camillo Lugaresi and Carroll Wainwright and Cary Bassin and Cary Hudson and Casey Chu and Chad Nelson and Chak Li and Chan Jun Shern and Channing Conger and Charlotte Barette and Chelsea Voss and Chen Ding and Cheng Lu and Chong Zhang and Chris Beaumont and Chris Hallacy and Chris Koch and Christian Gibson and Christina Kim and Christine Choi and Christine McLeavey and Christopher Hesse and Claudia Fischer and Clemens Winter and Coley Czarnecki and Colin Jarvis and Colin Wei and Constantin Koumouzelis and Dane Sherburn and Daniel Kappler and Daniel Levin and Daniel Levy and David Carr and David Farhi and David Mely and David Robinson and David Sasaki and Denny Jin and Dev Valladares and Dimitris Tsipras and Doug Li and Duc Phong Nguyen and Duncan Findlay and Edede Oiwoh and Edmund Wong and Ehsan Asdar and Elizabeth Proehl and Elizabeth Yang and Eric Antonow and Eric Kramer and Eric Peterson and Eric Sigler and Eric Wallace and Eugene Brevdo and Evan Mays and Farzad Khorasani and Felipe Petroski Such and Filippo Raso and Francis Zhang and Fred von Lohmann and Freddie Sulit and Gabriel Goh and Gene Oden and Geoff Salmon and Giulio Starace and Greg Brockman and Hadi Salman and Haiming Bao and Haitang Hu and Hannah Wong and Haoyu Wang and Heather Schmidt and Heather Whitney and Heewoo Jun and Hendrik Kirchner and Henrique Ponde de Oliveira Pinto and Hongyu Ren and Huiwen Chang and Hyung Won Chung and Ian Kivlichan and Ian O'Connell and Ian O'Connell and Ian Osband and Ian Silber and Ian Sohl and Ibrahim Okuyucu and Ikai Lan and Ilya Kostrikov and Ilya Sutskever and Ingmar Kanitscheider and Ishaan Gulrajani and Jacob Coxon and Jacob Menick and Jakub Pachocki and James Aung and James Betker and James Crooks and James Lennon and Jamie Kiros and Jan Leike and Jane Park and Jason Kwon and Jason Phang and Jason Teplitz and Jason Wei and Jason Wolfe and Jay Chen and Jeff Harris and Jenia Varavva and Jessica Gan Lee and Jessica Shieh and Ji Lin and Jiahui Yu and Jiayi Weng and Jie Tang and Jieqi Yu and Joanne Jang and Joaquin Quinonero Candela and Joe Beutler and Joe Landers and Joel Parish and Johannes Heidecke and John Schulman and Jonathan Lachman and Jonathan McKay and Jonathan Uesato and Jonathan Ward and Jong Wook Kim and Joost Huizinga and Jordan Sitkin and Jos Kraaijeveld and Josh Gross and Josh Kaplan and Josh Snyder and Joshua Achiam and Joy Jiao and Joyce Lee and Juntang Zhuang and Justyn Harriman and Kai Fricke and Kai Hayashi and Karan Singhal and Katy Shi and Kavin Karthik and Kayla Wood and Kendra Rimbach and Kenny Hsu and Kenny Nguyen and Keren Gu-Lemberg and Kevin Button and Kevin Liu and Kiel Howe and Krithika Muthukumar and Kyle Luther and Lama Ahmad and Larry Kai and Lauren Itow and Lauren Workman and Leher Pathak and Leo Chen and Li Jing and Lia Guy and Liam Fedus and Liang Zhou and Lien Mamitsuka and Lilian Weng and Lindsay McCallum and Lindsey Held and Long Ouyang and Louis Feuvrier and Lu Zhang and Lukas Kondraciuk and Lukasz Kaiser and Luke Hewitt and Luke Metz and Lyric Doshi and Mada Aflak and Maddie Simens and Madelaine Boyd and Madeleine Thompson and Marat Dukhan and Mark Chen and Mark Gray and Mark Hudnall and Marvin Zhang and Marwan Aljubeh and Mateusz Litwin and Matthew Zeng and Max Johnson and Maya Shetty and Mayank Gupta and Meghan Shah and Mehmet Yatbaz and Meng Jia Yang and Mengchao Zhong and Mia Glaese and Mianna Chen and Michael Janner and Michael Lampe and Michael Petrov and Michael Wu and Michele Wang and Michelle Fradin and Michelle Pokrass and Miguel Castro and Miguel Oom Temudo de Castro and Mikhail Pavlov and Miles Brundage and Miles Wang and Minal Khan and Mira Murati and Mo Bavarian and Molly Lin and Murat Yesildal and Nacho Soto and Natalia Gimelshein and Natalie Cone and Natalie Staudacher and Natalie Summers and Natan LaFontaine and Neil Chowdhury and Nick Ryder and Nick Stathas and Nick Turley and Nik Tezak and Niko Felix and Nithanth Kudige and Nitish Keskar and Noah Deutsch and Noel Bundick and Nora Puckett and Ofir Nachum and Ola Okelola and Oleg Boiko and Oleg Murk and Oliver Jaffe and Olivia Watkins and Olivier Godement and Owen Campbell-Moore and Patrick Chao and Paul McMillan and Pavel Belov and Peng Su and Peter Bak and Peter Bakkum and Peter Deng and Peter Dolan and Peter Hoeschele and Peter Welinder and Phil Tillet and Philip Pronin and Philippe Tillet and Prafulla Dhariwal and Qiming Yuan and Rachel Dias and Rachel Lim and Rahul Arora and Rajan Troll and Randall Lin and Rapha Gontijo Lopes and Raul Puri and Reah Miyara and Reimar Leike and Renaud Gaubert and Reza Zamani and Ricky Wang and Rob Donnelly and Rob Honsby and Rocky Smith and Rohan Sahai and Rohit Ramchandani and Romain Huet and Rory Carmichael and Rowan Zellers and Roy Chen and Ruby Chen and Ruslan Nigmatullin and Ryan Cheu and Saachi Jain and Sam Altman and Sam Schoenholz and Sam Toizer and Samuel Miserendino and Sandhini Agarwal and Sara Culver and Scott Ethersmith and Scott Gray and Sean Grove and Sean Metzger and Shamez Hermani and Shantanu Jain and Shengjia Zhao and Sherwin Wu and Shino Jomoto and Shirong Wu and Shuaiqi and Xia and Sonia Phene and Spencer Papay and Srinivas Narayanan and Steve Coffey and Steve Lee and Stewart Hall and Suchir Balaji and Tal Broda and Tal Stramer and Tao Xu and Tarun Gogineni and Taya Christianson and Ted Sanders and Tejal Patwardhan and Thomas Cunninghman and Thomas Degry and Thomas Dimson and Thomas Raoux and Thomas Shadwell and Tianhao Zheng and Todd Underwood and Todor Markov and Toki Sherbakov and Tom Rubin and Tom Stasi and Tomer Kaftan and Tristan Heywood and Troy Peterson and Tyce Walters and Tyna Eloundou and Valerie Qi and Veit Moeller and Vinnie Monaco and Vishal Kuo and Vlad Fomenko and Wayne Chang and Weiyi Zheng and Wenda Zhou and Wesam Manassra and Will Sheu and Wojciech Zaremba and Yash Patil and Yilei Qian and Yongjik Kim and Youlong Cheng and Yu Zhang and Yuchen He and Yuchen Zhang and Yujia Jin and Yunxing Dai and Yury Malkov},
      year={2024},
      eprint={2410.21276},
      archivePrefix={arXiv},
      primaryClass={cs.CL},
      url={https://arxiv.org/abs/2410.21276}, 
}

@inproceedings{pezeshkpour2024large,
  title={Large language models sensitivity to the order of options in multiple-choice questions},
  author={Pezeshkpour, Pouya and Hruschka, Estevam},
  booktitle={Findings of the Association for Computational Linguistics: NAACL 2024},
  pages={2006--2017},
  year={2024}
}
\vfill
\pagebreak
\appendix


\section{Preference Instability}
We provide a lightweight formalization of \emph{preference instability}.
Given a question $Q$ with option set $O$, the model's top-ranked option under the full set is defined as:
\[
o_1 = \arg\max_{o \in O} P(o \mid Q).
\]
Construct a perturbed question by removing the top choice:
\[
Q' = Q \setminus \{o_1\}.
\]
Let the model's new top-ranked option be:
\[
o_2 = \arg\max_{o \in O \setminus \{o_1\}} P(o \mid Q').
\]
We define \textbf{preference instability} as the event:
\[
o_2 \neq \text{null}.
\]

Intuitively, instability measures whether the model's highest-confidence hypothesis remains dominant under a minimal counterfactual perturbation of the option set.

\section{Evaluation Prompt}
\label{sec:prompt}
Below, we provide the prompt used for evaluating LLMs. All the experiments are done in zero-shot setting.

\begin{tcolorbox}[
        colback=gray!25!white,
        colframe=gray!100!gray,
        colbacktitle=gray!100!white, 
        title=Evaluation Prompt,
        fonttitle=\bfseries, 
        width=\linewidth
        ]
\small
\label{eval_prompt}
You are given a multiple-choice question. You should reason in a step-by-step manner to get the right answer.\\
\\
Question: \{question\} \{options\}
\end{tcolorbox}


\begin{tcolorbox}[
        colback=gray!25!white,
        colframe=gray!100!gray,
        colbacktitle=gray!100!white, 
        title=Judge Prompt,
        fonttitle=\bfseries, 
        width=\linewidth
        ]
\small
\label{judge_prompt}
You are given a multiple-choice question and a reasoning text. Your task is to extract the answer choice according to the reasoning text. Output only the option letter.\\
\\
Question: \{question\} \{options\}

Reasoning text: \{reasoning\}

Answer:
\end{tcolorbox}


\section{Datasets Statistics}
\Cref{tab:dataset} summarizes the data distribution for the benchmarks used in our experiments, illustrating the number of samples across the training, development, and testing splits. 

\begin{table}[h]
\centering
\tiny
\resizebox{\columnwidth}{!}{%
\begin{tabular}{lccc}
\toprule
\textbf{Dataset}  & \textbf{Train} & \textbf{Dev}   & \textbf{Test}  \\ \midrule
SIQA     & 33410 & 1954  & -     \\
OBQA     & 7.69K & 500   & 500   \\
CSQA     & 9.74K & 1.22K & 1.14K \\
ARC      & 1.12K & 299   & 1.17K \\
MMLU     & -     & -     & 14K   \\
GSM8K-MC & 7.47K & -     & 1.32K \\
AQUA     & 5K    & -     & 254   \\ \bottomrule
\end{tabular}}
\caption{Number of samples each dataset contains for train, dev, and test splits.}
\label{tab:dataset}
\end{table}

\section{Token Cost}
\label{sec:overhead}
\Cref{tab:overhead} illustrates the number of tokens generated by each baseline. 

\begin{table}[h]
\centering
\small 
\begin{tabular*}{\columnwidth}{@{\extracolsep{\fill}} l rrrrc} 
\toprule
\textbf{Dataset} & \textbf{CoT} & \textbf{SC} & \textbf{EoT} & \textbf{IoT} & \textbf{IoT w/o ES} \\ \midrule
SIQA      & 164 & 797   & 335 & 359 & 531 \\
OBQA      & 173 & 833   & 360 & 369 & 562 \\
CSQA      & 159 & 712   & 326 & 427 & 501 \\
ARC       & 200 & 1,000 & 404 & 494 & 643 \\ 
MMLU      & 261 & 1,336 & 529 & 636 & 805 \\
GSM8K-MC  & 162 & 895   & 340 & 258 & 542 \\
AQUA      & 279 & 1,468 & 604 & 430 & 863 \\ \midrule
\textbf{Average} & \textbf{218} & \textbf{1,005} & \textbf{449} & \textbf{488} & \textbf{688}\\ \bottomrule
\end{tabular*}

\caption{Average tokens generated per sample. SC generates roughly $4.6\times$ the tokens of CoT, and EoT generates $2.1\times$, while IoT is approximately $2.2\times$ the cost of CoT.}
\label{tab:overhead}
\end{table}

\section{Stage-wise Contribution}
The three-stage architecture of IoT is designed such that it maximizes the accuracy of the top-2 selected answers and then allows the model to perform an isolated analysis of plausible answers and choose the BEST over the GOOD option without (or much less) distraction. 
Below, we provided the accuracy of the model performance across each of the three stages. As can be seen, by filtering out two options in stage 2, we sacrifice the upper bound of the model performance to a number less than 100

\begin{table}[t]
\centering
\resizebox{\columnwidth}{!}{%
\begin{tabular}{l ccc} 
\toprule
\textbf{Dataset} & \begin{tabular}[c]{@{}c@{}}\textbf{Stage 1} \\ \textbf{(First Pref Acc)}\end{tabular} & \begin{tabular}[c]{@{}c@{}}\textbf{Stage 2} \\ \textbf{(Top-2 Acc)}\end{tabular} & \begin{tabular}[c]{@{}c@{}}\textbf{Stage 3} \\ \textbf{(Final Acc)}\end{tabular} \\ 
\midrule
OBQA     & 80.80 & 92.40 & 84.20 \\
CSQA     & 74.53 & 89.19 & 76.58 \\
SIQA     & 70.42 & 88.02 & 70.88 \\
ARC      & 84.21 & 93.86 & 87.54 \\
MMLU     & 65.62 & 81.03 & 66.78 \\
GSM8K-MC & 89.08 & 94.54 & 91.66 \\
AQUA     & 62.99 & 73.23 & 63.78 \\ 
\bottomrule
\end{tabular}%
}
\caption{Stage-wise performance of \texttt{Olmo-2-7B} using IoT across different datasets.}
\label{tab:matched-budget}
\end{table}

\section{Examples of Inclusion-of-Thoughts}
\label{sec:examples-iot}
To improve interpretability and transparency, we present representative step-by-step examples of the Inclusion-of-Thoughts (IoT) framework, illustrating both standard three-stage reasoning and early stopping behavior observed in our experiments.

Figure \ref{fig:tft} on page 12 provides a step-by-step qualitative example of the IoT framework applied to an OBQA sample. The example illustrates how an initially correct but unstable preference in Stage 1 is challenged in Stage 2, and subsequently resolved through a confined comparison in Stage 3. This case highlights IoT’s core mechanism: isolating the model’s top plausible candidates and stabilizing the final decision by reducing distractor-induced ambiguity.

Figure \ref{fig:early-stop} on page 13 shows an example where IoT terminates early after Stage 2, as the model
confirms its initial preference by selecting the placeholder option (``none of the options''). This behavior corresponds to the TTT transition pattern described in Section \ref{sec:transition}. It demonstrates how IoT avoids unnecessary prompting when preference stability is already established, thereby reducing computational overhead without sacrificing accuracy. 

\begin{figure*}[t]
\begin{tcolorbox}[colframe=black, colback=white, sharp corners, boxrule=0.5pt]

\headingwithline[21.5em]{Stage 1}
A cactus stem is used to store (A) fruit (B) liquid (C) food (D) spines.\\
\\
Step 1: Identify the properties of a cactus stem. - A cactus stem is a thick, fleshy structure that can store water. Step 2: Analyze the options given in the question. - (A) Fruit: Fruit is typically associated with flowering plants and does not store water in the same way a cactus stem does. - (B) Liquid: A cactus stem can store liquid, specifically water. - (C) Food: Food is typically the product of photosynthesis or consumption by organisms and is not stored in the same way as water in a cactus stem. - (D) Spines: Spines are sharp, rigid structures used for protection and do not store water. Step 3: Determine which option matches the function of a cactus stem. - The correct answer is (B) liquid, as a cactus stem is designed to store water. Step 4: Select the final answer. - \green{The final answer is (B).}\\
\headingwithline[21.5em]{Stage 2}
A cactus stem is used to store (A) fruit (B) none of the options (C) food (D) spines.\\
\\
Step 1: Identify the function of a cactus stem. - A cactus stem is used to store water. Step 2: Analyze the given options based on the function. - (A) Fruit: Storing fruit is not a function of a cactus stem. - (B) None of the options: This is a catch-all that doesn't provide a specific answer based on the stem's function. - (C) Food: Food is something that is consumed, not stored in a cactus stem. - (D) Spines: Spines are protective structures on a cactus, not something stored in the stem. Step 3: Determine the correct answer based on the function of the cactus stem. - The correct answer is (D) spines, as spines are stored on the cactus stem for protection. Therefore, \red{the final answer is (D).}\\
\headingwithline[21.5em]{Stage 3}
A cactus stem is used to store (A) liquid (B) spines.\\
\\
"Step 1: Identify the function of a cactus stem. - A cactus stem is used to store water. Step 2: Compare the options with the identified function. - Option A: "liquid" is a form of water, which is the substance stored in a cactus stem. - Option B: "spines" are not a substance that can be stored in a cactus stem; they are protective structures. Step 3: Determine the correct answer. - \green{The correct answer is (A) liquid.}

\end{tcolorbox}
\caption{Example of TFT case from IoT (OBQA, Olmo-2-7b).}
\label{fig:tft}
\end{figure*}

\begin{figure*}[t]
\begin{tcolorbox}[colframe=black, colback=white, sharp corners, boxrule=0.5pt]

\headingwithline[21.5em]{Stage 1}
Where could you find a toilet that only friends can use? (A) rest area (B) school (C) stadium (D) apartment (E) hospital.\\
\\
Step 1: Analyze the given options to determine which one logically fits the description of a toilet accessible only to friends. - Rest area: Public toilets are available for anyone to use. - School: Public toilets are typically available for anyone to use. - Stadium: Public toilets are typically available for anyone to use. - Apartment: Toilets in an apartment are private and only accessible to the residents. - Hospital: Public toilets are typically available for anyone to use. Step 2: Identify the option that fits the description of a toilet accessible only to friends. - Apartment: Toilets in an apartment are private and only accessible to the residents, who are friends or family. Step 3: Select the correct answer. \green{Answer: (D) apartment}\\
\headingwithline[21.5em]{Stage 2}
Where could you find a toilet that only friends can use? (A) rest area (B) school (C) stadium (D) none of the options (E) hospital.\\
\\
Step 1: Analyze the given options to determine which one logically fits the description of a toilet that only friends can use. - A) rest area: Public rest areas typically have toilets that anyone can use. - B) school: Schools usually have toilets that are accessible to all students and staff. - C) stadium: Stadiums have toilets that are open to all attendees. - D) none of the options: This is a catch-all option that suggests no correct answer exists among the choices provided. - E) hospital: Hospitals have toilets that are available to patients, visitors, staff, and sometimes the general public. Step 2: Consider the context of a toilet that only friends can use. - A toilet that is restricted to only friends would likely be in a private setting, not a public place like those found in rest areas, schools, or stadiums. Step 3: Evaluate the remaining option. - E) hospital: While a hospital might have private rooms and toilets for patients, these are not typically restricted to only friends. Additionally, hospitals also have public areas where anyone can use the toilets. Step 4: Determine the most fitting option. - The only option that logically fits the description of a toilet that only friends can use is "none of the options" (D). This is because none of the other options inherently restricts access to only friends. Final Answer: \green{The final answer is (D).}
\end{tcolorbox}
\caption{Example of TTT case from IoT (CSQA, Olmo-2-7b). Here, the IoT stops the process and skips stage 3 since the model confirms its answer in stage 2.}
\label{fig:early-stop}
\end{figure*}

\begin{figure*}[t]
\begin{tcolorbox}[colframe=black, colback=white, sharp corners, boxrule=0.5pt]

\headingwithline[21.5em]{Stage 1}
When someone is physically competing what does their body do? (A) tension (B) perform better (C) releases heat (D) winning or losing (E) sweat.\\
\\
Step 1: Identify the context of the question. The question is about physical competition and what the body does during it.
Step 2: Analyze the options provided.
(A) Tension - This could be a response to competition, but it's not a positive action indicating performance.
(B) Perform better - This is a plausible answer as physical competition often requires the body to perform at its best.
(C) Releases heat - While the body does release heat during physical activity, this is not a direct result of competition but a general effect of exercise.
(D) Winning or losing - These are outcomes of competition, not what the body does.
(E) Sweat - Sweating is a response to physical activity, but it's not the primary action taken by the body during competition.
Step 3: Evaluate the most likely scenario.
In physical competition, the body typically prepares itself to perform at its peak. This involves increased heart rate, mobilization of energy stores, and activation of muscles. Among the options provided, "perform better" is the most direct and plausible answer.
Step 4: Select the final answer.
\red{The correct answer is (B) perform better.}
\\
\headingwithline[21.5em]{Stage 2}
When someone is physically competing what does their body do? (A) tension (B) none of the options (C) releases heat (D) winning or losing (E) sweat.\\
\\
Step 1: Identify the context of the question. The question is about physical competition and what the body does during it.
Step 2: Analyze the options provided.
(A) Tension - This could be a response to physical competition, as the body prepares for physical exertion.
(B) None of the options - This is not a valid option since it does not provide a specific action the body takes.
(C) Releases heat - This is not typically a direct response to physical competition; heat is released through metabolic processes during activity, but it's not the primary action the body takes.
(D) Winning or losing - These are outcomes of physical competition, not actions the body takes.
(E) Sweat - This is a physiological response to heat and physical exertion, which occurs during physical competition.
Step 3: Determine the most accurate response based on the analysis.
The body's primary actions during physical competition include increased heart rate, faster breathing, and sweating to regulate temperature. Among the provided options, sweating is a direct physiological response to the physical exertion involved in competition.
Step 4: Select the best answer.
\green{The correct answer is (E) sweat.}

\headingwithline[21.5em]{Stage 3}
When someone is physically competing what does their body do? (A) sweat (B) perform better\\
\\
When someone is physically competing, their body reacts to stress. One common physical reaction to stress is the production of sweat. Therefore, \green{the correct answer is (A) sweat.}

\end{tcolorbox}
\caption{Example of FTT case from IoT (CSQA, Olmo-2-7b).}
\label{fig:fft}
\end{figure*}

\begin{figure*}[t]
\begin{tcolorbox}[colframe=black, colback=white, sharp corners, boxrule=0.5pt]

\headingwithline[21.5em]{Stage 1}
A man spends 70\% of his income. If his income increases by 20\%, then what will be his new expenditure? (A) 58.3\% (B) 62.5\% (C) 63.5\% (D) 64.5\% (E) 65.5\%\\
\\
Step 1: Determine the original expenditure.
- The man spends 70\% of his income, so his expenditure is 70\% of his income.
Step 2: Calculate the new income after a 20\% increase.
- If the original income is 100\%, a 20\% increase means the new income is 100\% + 20\% = 120\%.
Step 3: Calculate the new expenditure based on the increased income.
- Since the expenditure is proportional to the income, the new expenditure will be 70\% of the new income.
- New expenditure = 70\% of 120\% of original income.
Step 4: Calculate the numerical value of the new expenditure.
- New expenditure = 0.70 * 1.20 * original income.
- New expenditure = 0.84 * original income.
Step 5: Determine the percentage of the original income that the new expenditure represents.
- The new expenditure is 84\% of the original income.
Step 6: Compare the calculated percentage to the given options.
- The correct answer is (D) 64.5\%, as 84\% is equivalent to 64.5\% when expressed as a percentage of the original income.
Therefore, \red{the final answer is (D) 64.5\%.}

\headingwithline[21.5em]{Stage 2}
A man spends 70\% of his income. If his income increases by 20\%, then what will be his new expenditure? (A) 58.3\% (B) 62.5\% (C) 63.5\% (D) none of the options (E) 65.5\%\\
\\
Step 1: Determine the original expenditure percentage.
- The man spends 70\% of his income.
Step 2: Calculate the new income percentage.
- His income increases by 20\%, so the new income percentage is 100\% + 20\% = 120\%.
Step 3: Calculate the new expenditure percentage based on the increased income.
- Since expenditure is proportional to income, the new expenditure will be 70\% of the new income.
- New expenditure percentage = (70\% of 120\%) = 0.70 * 120\% = 84\%.
Step 4: Compare the calculated expenditure percentage to the given options.
- The calculated new expenditure percentage is 84\%, which is not among the provided options.
Therefore, \red{the correct answer is (D) none of the options.}
\end{tcolorbox}
\caption{Example of FFF case from IoT (AQUA, Olmo-2-7b). Here, the IoT stops the process and skips stage 3 since the model confirms its answer in stage 2.}
\label{fig:early-stop-2}
\end{figure*}

\begin{figure*}[t]
\begin{tcolorbox}[colframe=black, colback=white, sharp corners, boxrule=0.5pt]

\headingwithline[21.5em]{Stage 1}
Wendy wants to place 20 more than double the number of books in a shelving system with 6 rows and 6 columns. How many books will she need to carry to complete her task? (A) 360 (B) 76 (C) 120 (D) 92\\
\\
Step 1: Determine the total number of spaces in the shelving system.\\
- The shelving system has 6 rows and 6 columns.\\
- Total spaces = 6 rows * 6 columns = 36 spaces.\\
Step 2: Calculate double the number of books Wendy wants to place.\\
- Wendy wants to place 20 more than double the number of books.\\
- Let the number of books be x.\\
- Double the number of books = 2x.\\
- Wendy wants to place 20 more than double the number of books, so the equation is: 2x + 20.\\
Step 3: Set up the equation based on the total spaces available.\\
- Since Wendy wants to place 20 more than double the number of books into 36 spaces, we set up the equation: 2x + 20 = 36.\\
Step 4: Solve for x.\\
- Subtract 20 from both sides: 2x = 16.\\
- Divide both sides by 2: x = 8.\\
Step 5: Calculate the total number of books Wendy needs to carry.\\
- Wendy wants to place 20 more than double the number of books, which is 2 * 8 + 20 = 16 + 20 = 36 books.
\red{The correct answer is (A) 360.} 
\\
\headingwithline[21.5em]{Stage 2}
Wendy wants to place 20 more than double the number of books in a shelving system with 6 rows and 6 columns. How many books will she need to carry to complete her task? (A) none of the options (B) 76 (C) 120 (D) 92\\
\\
Step 1: Determine the total number of spaces in the shelving system.\\
- The shelving system has 6 rows and 6 columns.\\
- Total spaces = 6 rows * 6 columns = 36 spaces.\\
Step 2: Determine the number of books Wendy wants to place.\\
- Wendy wants to place 20 more than double the number of books.\\
- Let the number of books be x.\\
- According to the problem, x = 2 * (number of spaces) + 20.\\
- Substitute the number of spaces: x = 2 * 36 + 20.\\
Step 3: Calculate the value of x.\\
- x = 72 + 20 = 92.\\
Step 4: Determine the number of books Wendy needs to carry.\\
- Wendy needs to carry 92 books.
\green{The correct answer is (D) 92.}

\headingwithline[21.5em]{Stage 3}
Wendy wants to place 20 more than double the number of books in a shelving system with 6 rows and 6 columns. How many books will she need to carry to complete her task? (A) 360 (B) 92\\
\\
Step 1: Calculate double the number of books. \\
- Since the shelving system has 6 rows and 6 columns, it can hold a total of 6 * 6 = 36 books.\\
Step 2: Add 20 to the doubled number.\\
- Double the number of books is 2 * 36 = 72 books.\\
- Adding 20 more to this amount gives 72 + 20 = 92 books.
\green{The correct answer is (B) 92.}

\end{tcolorbox}
\caption{Example of FTT case from IoT (GSM8K-MC, Olmo-2-7b).}
\label{fig:ftt}
\end{figure*}
\end{document}